\documentclass{article}

\usepackage{arxiv}

\usepackage[utf8]{inputenc} 
\usepackage[T1]{fontenc}    
\usepackage{hyperref}       
\usepackage{url}            
\usepackage{booktabs}       
\usepackage{amsfonts}       
\usepackage{nicefrac}       
\usepackage{microtype}      
\usepackage{lipsum}		
\usepackage{graphicx}
\usepackage[numbers]{natbib}
\usepackage{doi}

\usepackage{amssymb}
\usepackage{amsmath,amsfonts}
\usepackage{multirow}
\usepackage{subcaption}
\usepackage{algorithm,algpseudocode}
\newcommand{\RNum}[1]{\uppercase\expandafter{\romannumeral #1\relax}}
\usepackage[modulo]{lineno}
\usepackage{xcolor}
\usepackage{booktabs}
\usepackage{array}
\usepackage{soul}
\usepackage{float}

\newcommand{\thickbar}[1]{\mathbf{\bar{\text{$#1$}}}}

\title{MaskRenderer: 3D-Infused Multi-Mask Realistic Face Reenactment}

\author{ \href{https://orcid.org/0000-0002-9779-8754}{\includegraphics[scale=0.06]{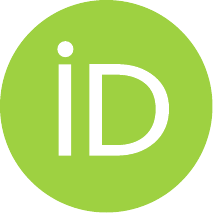}\hspace{1mm}Tina Behrouzi$^*$}\\
	Department of Computer Science\\
	University of Toronto\\
	\texttt{tina.behrouzi@mail.utoronto.ca} \\
	\And
	\href{https://orcid.org/0000-0003-0355-1150}{\includegraphics[scale=0.06]{orcid.pdf}\hspace{1mm}Atefeh Shahroudnejad$^*$}\\
	Amii (Alberta Machine Intelligence Institute)\\
	\texttt{atefeh.shahroudnejad@amii.ca} \\
	\And
	\href{https://orcid.org/0000-0003-3649-4703}{\includegraphics[scale=0.06]{orcid.pdf}\hspace{1mm}Payam Mousavi}\\
	Amii (Alberta Machine Intelligence Institute)\\
	\texttt{payam.mousavi@amii.ca} \\
}

\date{}

\hypersetup{
pdftitle={MaskRenderer},
pdfauthor={Tina Behrouzi, Atefeh Shahroudnejad, Payam Mousavi},
pdfkeywords={First keyword, Second keyword, More},
}
\newcommand\nnfootnote[1]{%
  \begin{NoHyper}
  \renewcommand\thefootnote{*}\footnote{#1}%
  \addtocounter{footnote}{-1}%
  \end{NoHyper}
}

\begin{document}
\maketitle
\nnfootnote{These authors contributed equally.}
\begin{abstract}
We present a novel end-to-end identity-agnostic face reenactment system, MaskRenderer, that can generate realistic, high fidelity frames in real-time. 
Although recent face reenactment works have shown promising results, there are still significant challenges such as identity leakage and imitating mouth movements, especially for large pose changes and occluded faces. MaskRenderer tackles these problems by using (i) a 3DMM to model 3D face structure to better handle pose changes, occlusion, and mouth movements compared to 2D representations; (ii) a triplet loss function to embed the cross-reenactment during training for better identity preservation; and (iii) multi-scale occlusion, improving inpainting and restoring missing areas. Comprehensive quantitative and qualitative experiments conducted on the VoxCeleb1 test set, demonstrate that MaskRenderer outperforms state-of-the-art models on unseen faces, especially when the Source and Driving identities are very different.
\end{abstract}


\section{Introduction}
\begin{figure*}[!t]
		\centerline{\includegraphics[width=11cm]{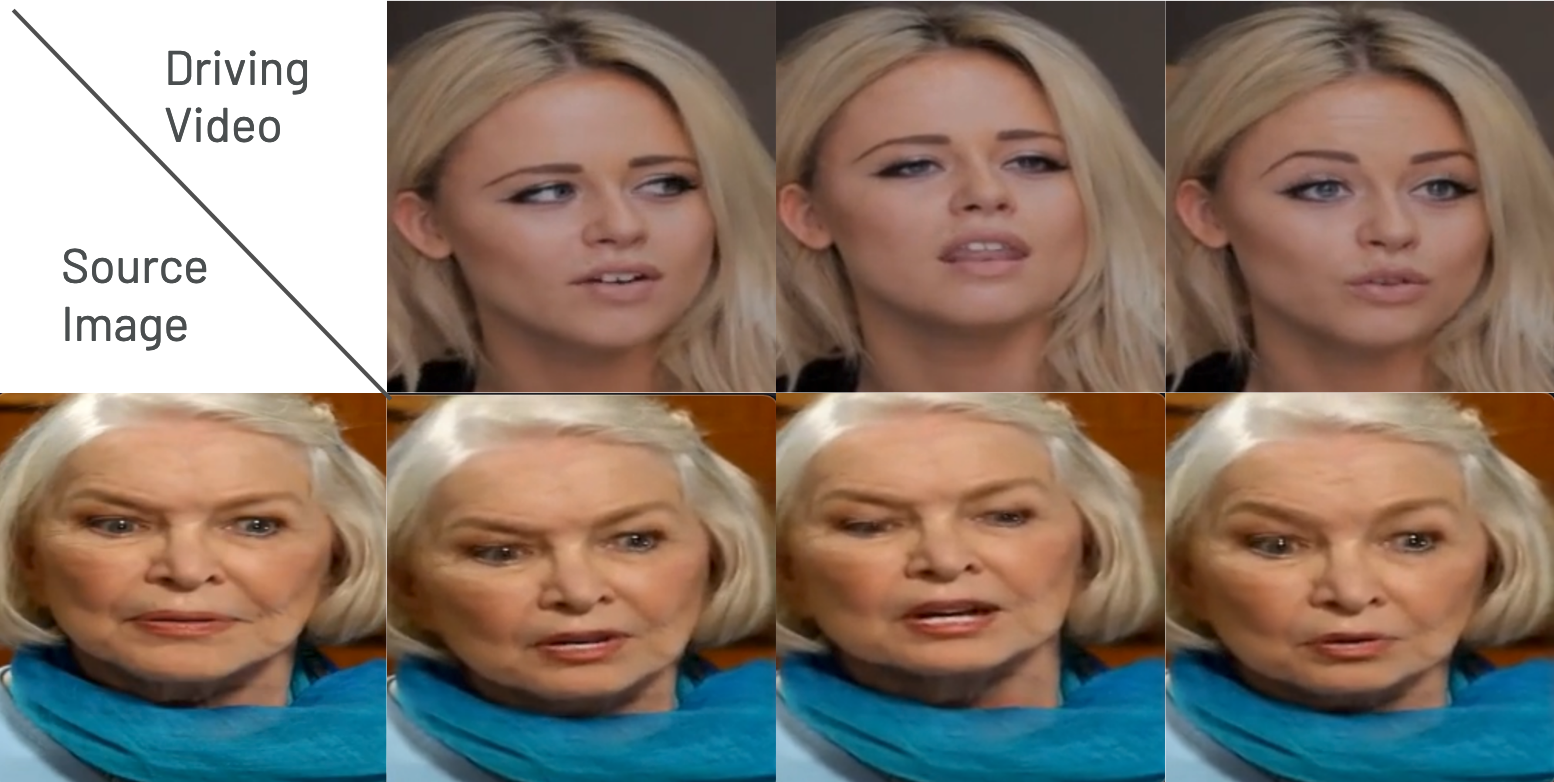}}
		\caption{This example face reenactment result is generated by MaskRenderer.}
		\label{fig:intro}
\end{figure*}

Identity-agnostic face reenactment is the process of generating sequential face images of a target person (i.e., Source), where the talking (i.e., Driving) person's pose and facial expression are transferred to the target person's image. In other words, the talking person controls the Source image and its pose and expression like a puppet (See Fig.~\ref{fig:intro}).

This well-established computer vision problem has a wide range of application areas, such as the film industry, teleconferencing, and virtual reality \cite{tolosana2020deepfakes}. However, high-fidelity and identity-agnostic face reenactment still remains challenging, particularly when the model should (i) preserve the Source's identity, (ii) generate a photo-realistic face, and (iii) ideally require only a single image of the Source. More precisely, the algorithm is required to preserve the face texture and shape of the Source, even in the presence of occlusion or any changes in the expression or pose of the Driving. 
We assume that only one photo is available from the Source, and the model is suitable for real-time applications. 

\subsection{Background}
In contrast to identity-agnostic face reenactment for which a single Source image is used, two other face reenactment categories use multiple frames (i.e., few-shot) or video footage (i.e., identity-specific). However, few-shot methods generally only work well when all Source images are extracted from the same video because photos of the same person at different times typically have different backgrounds, lighting conditions, and facial details (e.g., hair). That alteration in images can make identity preservation harder and worsen the cross-reenactment quality, compared to the single-image methods. Moreover, some few-shot methods such as Vid2Vid \cite{wang2019few} and MarioNETte \cite{ha2020marionette} are not real-time and have high computational costs.

Most face reenactment methods use one or more facial representations to accurately understand and capture face structure (i.e., identity), pose, and expression. 
Facial landmarks, a common representation, have been used to determine the pose and expression changes in Driving videos \cite{zhang2020freenet,hsu2022dual}. Sparse landmarks, trained in a supervised manner, are fixed points that cannot fully capture facial structure and features.
Moreover, a heavy dependence on supervised landmarks makes the model more sensitive to occlusion and inaccurate prediction. Recent works \cite{wang2021safa,wang2022latent,siarohin2019first} have shown that unsupervised facial features can better represent face motion for reenactment. Moreover, unsupervised facial features better depict pose changes than supervised key points. 

The vast majority of existing works~\cite{siarohin2019first,zhao2022thin} find facial representation and motion based only on a person's 2D image. However, these models do not capture 3D geometry from other views with different poses and expressions. 
To tackle this issue, a 3D Morphable Model (3DMM) \cite{deng2019accurate} can be used as a 3D representation to enhance the robustness of face reconstruction and improve mouth movements, particularly in the presence of occlusion and large pose changes. Furthermore, 3DMM models extract person-specific disentangled parameters such as pose, lighting, and identity.
The original 3DMM employs Principal Component Analysis (PCA) to find statistical properties of face shape. More recent 3DMMs \cite{fan2023landmark,feng2021learning} have included a light Neural Network or a GAN model to generate a full 3D representation of a face. 

Most 3DMM models \cite{fan2023landmark} are trained on 3D face scans that are extremely expensive to collect and, due to annotation costs, they have less variety in expression. To tackle this problem, some recent works, such as DECA \cite{feng2021learning}, can estimate accurate 3D faces based on images in the wild. However, these models usually require multiple views of the face to construct a detailed texture map that is robust to occlusion and pose changes.
Although 3DMM models have shown impressive results in generating a 3D face, they still suffer in recovering facial components, such as eyes, neck, teeth, facial detail (e.g., beard and makeup), and image background. To address these limitations, there are image inpainting methods, such as DE-GAN \cite{zhang2022gan}, that can help in recovering a face image. 

\subsection{Contributions}
This article introduces the MaskRenderer framework that reenacts the face of an unseen Source based on a sequence of Driving frames while preserving the Source identity. The proposed model can generate images in real-time. 
Our main contributions are as follows: 
 \begin{itemize}
 \item Inspired by the recent advance in 3D vision models, we incorporated Deng 3DMM \cite{deng2019accurate} into our model. Combining 3D head parameters and 2D motion transformation resulted in more accurate pose changes and iris and mouth movements. 
 \item Current face reenactment models do not perform cross-reenactment during training. Not considering different identities for Source and Driving during training causes under-performance at inference time when the Source and Driving images are from two different people. We designed a triplet loss to incorporate cross-reenactment loss during training, which improves identity preservation 
 where the Source and Driving facial shapes are completely different.
 \item We used multi-scale occlusion in the generator for better inpainting of unwanted or missing areas.
 \item MaskRenderer outperformed state-of-the-art methods on the VoxCeleb1 benchmark in terms of identity preservation, transferring pose and expression, and mouth movements.
 \end{itemize}

The remaineder of the article is as follows: In Section \ref{sec:Related}, recent related works on face reenactment are discussed. The framework of our proposed method and the detailed structure are explained in Section \ref{sec:Method}. Section \ref{sec:Eval} summarizes the datasets, inference requirements, evaluation metrics, and training paradigm. In section \ref{sec:Result}, MaskRenderer is compared with baselines and other state-of-the-art methods. Section \ref{sec:Ablation} contains empirical ablation studies to evaluate and justify our algorithmic design decisions. Finally, Section \ref{sec:Conclude} concludes and discusses the future directions.

\section{Related Works}
\label{sec:Related}
In this section, we review existing face reenactment methods and categorize them based on the three types of face representation. 
 
\subsection{3D-based models}
3D-based models leverage 3D facial geometry to improve upon 2D techniques, better controlling different expressions and poses. 
The notable challenges in 3D-based models are (i) achieving high-fidelity identity preservation and looking realistic instead of like a 3D avatar, (ii) some tasks, such as hair generation, is more problematic in 3D, and (iii) the quality of 3D animatable head reconstruction typically depends on the extracted geometry, texture rendering, model generalization, and disentanglement between identity, pose, and expression. 
Earlier works \cite{thies2016face2face,kim2018deep,thies2018headon} transfer 3D facial parameters from Driving to Source and generate the reenactment video by rendering the input. The primary limitation of these works is that they all need to be trained on a multi-view video of a specific person.
Many recent agnostic works~\cite{wang2021safa,ren2021pirenderer,yin2022styleheat,doukas2021headgan,lin20223d,yangface2face} incorporate a pre-trained 3DMM
into their structure to model the shape, texture, expression, and pose of faces.
PIRenderer~\cite{ren2021pirenderer} first maps 3DMM parameters to a latent space and then uses it to guide the motion network and the generator.
Yao et al.~\cite{yao2020mesh} predicts 3DMM parameters to reconstruct 3D face meshes of Source and Driving. Then, they apply graph Convolutional operations on these meshes to extract motion features to guide the motion estimation network.

\subsection{Landmark-based models}
In this class of model, facial landmarks are used as a prior condition to guide the generator by representing pose and expression. The main challenge in landmark-based models is preserving identity due to the person-specific identity information in facial landmarks.
Zakharov et al. \cite{zakharov2019few} proposed NeuralHead as a few-shot face reenactment that generates reenactment directly from landmarks by adopting a meta-learning strategy and fine-tuning for unseen identities. A personalized pose-invariant embedding vector is created based on different sets of frames and is used to predict AdaIN parameters~\cite{huang2017arbitrary} in the generator for appearance transfer. NeuralHead was later extended~\cite{burkov2020neural} for better cross-identity reenactment by using latent pose vectors instead of directly employing landmarks. 

Another few-shot work to address the identity preservation problem is MarioNETte~\cite{ha2020marionette}. It introduces a landmark transformer to adapt facial points between Source and Driving and uses an image attention block for better appearance style transfer. The idea of a landmark transformer was later used in multiple works \cite{zhang2020freenet,liu2021li,sun2022landmarkgan}.
To improve the visual quality of results, others~\cite{zhao2021sparse,yao2021one} consider facial components landmarks separately, in addition to the facial landmarks in the reenactment process. 
Hsu et al.~\cite{hsu2022dual} employ a 3D landmark detector, rather than 2D, to cope with large pose changes. 

\subsection{Motion-based models}
Motion-based models are composed of two sub-networks: (i) a motion estimation network to predict a dense motion field from Source to Driving that is used for warping the Source and (ii) a generative network to generate the final reenactment result using the warped Source. Siarohin et al.~\cite{siarohin2019monkeynet} first proposed MonkeyNet, which encodes motion by learning self-supervised keypoints. Then in their follow-up work, named First-Order Motion Model (FOMM)~\cite{siarohin2019first}, they significantly improved the results' appearance, particularly for more complex motions, by considering local affine transformations instead of rigid motions. Many subsequent works, including ours, have drawn inspiration from the FOMM's performance and structure. 
Zhao et al.~\cite{zhao2022thin} reformulated FOMM for better transferring the motion when there is a large pose difference between Source and Driving. They employ Thin-Plate Spline transformations along with the affine transformation for the dense motion estimation in the training path.

Another work tackling the large pose changes between Source and Driving is proposed by Liu et al.~\cite{liu2021self}. They adopted Neural-ODE to differentially refine the initial keypoints and model the dynamics of motion deformation.  
Tripathy et al.~\cite{tripathy2022single} use a keypoint transformer to stabilize keypoints in FOMM for better cross-identity reenactment.
To address the limitations of 2D representations in motion changes and cross-identity issues, Hong et al.~\cite{hong2022depth} proposed estimating facial depth maps from Source and Driving and using them as 3D facial geometries to guide the keypoint estimator in FOMM. These depth maps are also used to refine the motion field for more detailed and accurate reenactment. Wang et al.~\cite{wang2021one} also extended FOMM by predicting canonical 3D keypoints instead of 2D to support large pose changes.

In a recent work, LIA~\cite{wang2022latent} uses StyleGAN2~\cite{karras2020analyzing} and sparse coding to estimate the motion from Source to Driving without keypoints prediction. Therefore, the quality of reenactment results is not dependent on the initial Driving frame.
Other methods~\cite{siarohin2021motion,shalev2022image} extract regions or masks instead of keypoints to separate the foreground from the background for improving shapes capturing. Hence, these methods are more suitable for articulated objects like human bodies.
Our proposed model comprises both motion-based and 3D-based models resulting in high-quality, high-fidelity, and photo-realistic face reenactment.

\section{MaskRenderer: Methodology and Approach}
\label{sec:Method}
\begin{figure*}[!t]
		\centerline{\includegraphics[trim=100 0 100 0, width=13cm]{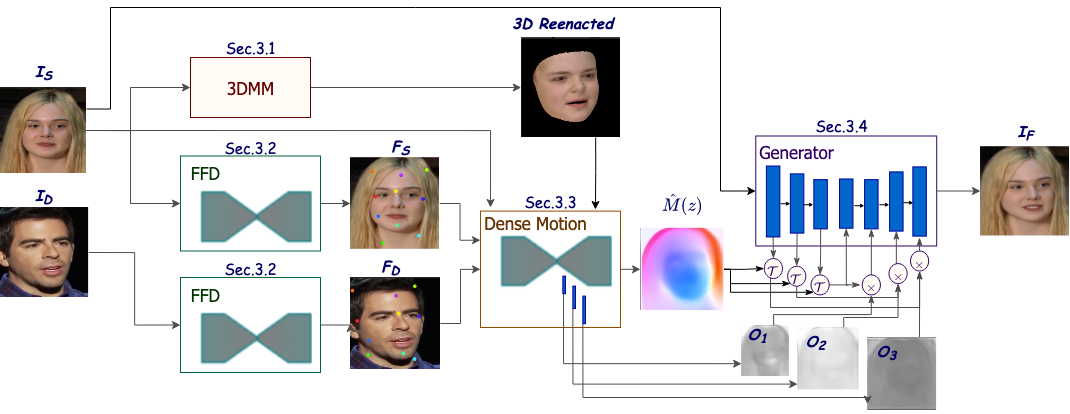}}
		\caption{The architecture of the proposed model}
		\label{fig:overview}
\end{figure*}

In this section, we explain the detailed structure of the MaskRenderer. 
Fig.~\ref{fig:overview} shows a high-level schematic. The network is divided into four parts: (i) the 3DMM module generates disentangled Source's facial shape and Driving's pose and expression. (ii) the Facial Features Detector (FFD) finds representative facial characteristics. Then, both the FFD features and 3DMM parameters are passed to (iii), the Dense Motion network, to find the optical flow between the Source and Driving images. In (iv), the Multi-scale Occlusion Masks are constructed to represent the image's missing parts. The generator takes the occlusion masks and dense motion matrix to create the reenacted result. MaskRenderer is a GAN-based model that uses a Generative Adversarial Network (GAN) architecture to generate realistic reenacted images from a given source image. 
The final part of this section is devoted to discussing our novel use of triplet loss among other loss terms. 

\subsection{3DMM}
Given the Source image $I_S$, the 3DMM objective is to reproduce the 3D shape of $I_S$, 
which is consistent with various image viewpoints. 
The original 3DMM is defined as Eq.~\eqref{equ:3DMM_S}, where $S$ represents the 3D face shape.  
 $\thickbar{S}$ is the mean face shape value. $\alpha_{id}$ and $\alpha_{exp}$ are identity and expression coefficients, and finally $A_{id}$ and $A_{exp}$ are the identity and expression PCA bases, respectively. Further explanation can be found in the supplementary material and the 3DMM paper \cite{cao2013facewarehouse}.
\begin{equation}
    S = \thickbar{S} + A_{id}\alpha_{id} + A_{exp}\alpha_{exp}
    \label{equ:3DMM_S}
\end{equation}
\begin{equation}
    T = \thickbar{T} + A_{tex}\alpha_{tex}
    \label{equ:3DMM_T}
\end{equation}

The skin texture $T$ is combined with the face shape $S$, defined in Eq.~\eqref{equ:3DMM_T}, where the texture coefficient $\alpha_{tex}$ adjusts the mean texture value $\thickbar{T}$ and texture PCA base $A_{tex}$. The input image only changes the coefficients ($\alpha_{id} \in \mathbb{R}^{80}$, $\alpha_{exp} \in \mathbb{R}^{64}$, $\alpha_{tex} \in \mathbb{R}^{80}$). The PCA bases and mean values ($A_{id}$, $A_{exp}$, $A_{tex}$, $\thickbar{S}$, $\thickbar{T}$) are set to be consistent with prior work~\cite{cao2013facewarehouse}.

We employ the pre-trained Deep3DFace model \cite{deng2019accurate} as the 3DMM model as this model finds an explicit texture's mesh structure from a single-view image and can capture major changes in the head and mouth area. Since the 3DMM's features are disentangled, the pose and expression can be taken from the Driving frame $I_D$, while other parameters are derived from the Source image $I_S$.
The 3D reenacted face and motion field between the Source and Driving face vertexes are passed to the Dense Motion network. 

Although the reenacted 3D face shape is stable to different poses, expressions, and lighting, the final texture is not detailed and does not include regions such as teeth, hair, and inner eye parts. We address this issue in the later steps, bottom right of Fig. \ref{fig:overview}, where the skin texture and facial details are reconstructed in 2D space. 

\subsection{Facial Feature Detection}

\begin{figure*}[!t]
		\centerline{\includegraphics[trim=50 0 50 0,width=13cm]{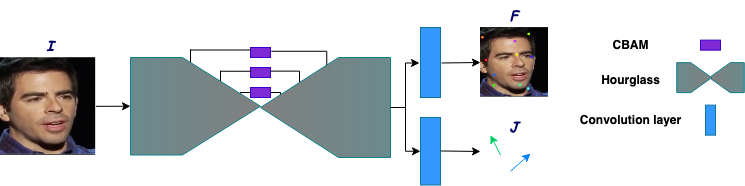}}
		\caption{In the FFD module, the attention module is applied to the skip connections' features}
		\label{fig:hourglass}
\end{figure*}

The Facial Feature Detection (FFD) structure is inspired by FOMM's keypoints detector network~\cite{siarohin2019first}. The FFD structure and the corresponding equivariance losses are designed to ensure that facial features\footnote{In this article, we use the term ``feature'' instead of ``keypoint'' because the unsupervised features do not directly represent facial landmarks.} are presentable on an input face. 
As shown in Fig.~\ref{fig:hourglass}, the input image $I \in \mathbb{R}^{256\times256\times3}$ (from either the Source or the Driver) is passed through the hourglass network to create a heatmap with the shape of $(K\times64\times64)$, where $K$ is the spatial feature shape. The heatmap is passed via two parallel convolution layers to create the feature locations in 2D coordinate $F \in \mathbb{R}^{K \times 2}$ and motion matrix $J \in \mathbb{R}^{K \times 2 \times 2}$.

The facial features are used to identify the sparse motion and dense motion, subsection \ref{subsec:dense}. The sparse motion matrix $M_{k}(z)$ between Source and Driving images is defined as:
\begin{equation}
    M_{k}(z) = F_{S_{k}} + (J_{S_{k}}J_{D_{k}}^{-1}) (z - F_{D_{k}}),
    \label{equ:FFD}
\end{equation}

where $M_{k}(z)$ is calculated at coordinate positions $z$ near each feature point for each feature row $k \in \{1,\dotsc, K\}$. The subscripts $S$ and $D$ denote whether features belong to the Source or Driving image. The learnable $J_S J_D^{-1}$ matrix control the impact of changes in $F_D$ on $F_S$.

We have included the attention mechanism in the hourglass model by adding Convolutional Block Attention Module (CBAM)~\cite{woo2018cbam} blocks. The new hourglass can capture the gradual changes between $F_S$ and $F_D$ using channel and spatial attention. 
The attention blocks also help FFD to be robust against occlusion, face rotation, and changes in illumination by considering the positional relationship between features and only influential information.

\subsection{Dense Motion Network}
\label{subsec:dense}
The objective of the Dense Motion network is to find the optical flow between $I_S$ and $I_D$, transferring the Driving's pose and expression into the Source image. However, deriving dense motion between $I_S$ and $I_D$ is very challenging due to the small $K$ number of unsupervised features $F$. Hence, we incorporated 3DMM-based parameters to better address motions such as eye blinking, head movement, and mouth opening. The sparse motion $M_k(z)$ is concatenated with the motion field $M_{3D}(z)$ derived from the 3DMM module to better represent the flow changes from the Driving to the Source's face.

To estimate the final dense motion $\hat{M}(z)$, $K+2$ 
dense heatmaps' masks $H$ are required to combine static points, 3D motions, and the $K$ local features (Eq.~\ref{equ:dense_motion}). To generate $H_{0}$, $H_{3D}$, and $H_{k}$ matrices, another attention hourglass network is applied respectively to
the sparse transformation of the Source image, the reenacted 3D face, and to the features. 
\begin{equation}
    \hat{M}(z) = H_{0}z + H_{3D} M_{3D}(z) + \sum_{k=1}^{K} H_{k}M_k(z)
    \label{equ:dense_motion}
\end{equation}

\subsection{Multi-Scale Occlusion Masks \& Generator}
Occlusion masks address the pixel locations that are missing or occluded in the Source image by inpainting them during generation. However, different feature maps in the Dense Motion module's decoder have different relative importance. The high-level feature map represents facial details such as face and color, while the low-level feature map focuses on coarse features such as pose and face shape. Accordingly, we define an occlusion mask $O_i \in [0,1]$ for the feature map in layer $i$, as shown bottom right part of the Fig. \ref{fig:overview}. The $O_i$ is estimated using a parallel convolutional layer with a sigmoid activation function on the output of the attention hourglass's decoder layer $i$ in the Dense Motion module.

To generate a photo-realistic reenacted 2D image, a generator with an encoder-decoder structure is designed. The encoder extracts Source features for different scales, $F_{Enc_{S, i}}$, where $i$ is the $i^{th}$ down-sampled layer. The decoder employs per-pixel mapping derived from the dense motion module to warp the Source features and up-sample $F_{Enc_{S}}$ to the original image shape.  

After warping $F_{Enc_{S, i}}$ based on the Driving's pose and expression,
the missing pixels should be inpainted.  
 For this complex inpainting, we apply the occlusion mask $O_i$ to the warped features, $\mathcal{T}(F_{Enc_{S, i}},\hat{M}(z))$, to locate the hidden regions that require reconstruction. The decoder's output at layer $i$ is concatenated with $O_i \times \mathcal{T}(F_{Enc_{S, i}},\hat{M}(z))$ and passed to the next up-sample layer to generate the final reenacted image $I_F$.

\subsection{Training Losses}
The model is trained in a self-supervised fashion. All the training losses, except the triplet loss, are defined based on self-reenactment, where $I_S$ and $I_D$ are two frames of the same person in a video. The triplet loss is based on cross-reenactment and uses identity soft labels derived from the VGGFace2 \cite{cao2018vggface2}. 
MaskRenderer is trained end-to-end, and all the losses listed below are summed and minimized. 

\textbf{\textit{Triplet loss $L_T$:}}
To encourage the network to generate identity-consistent images robust to changes in pose and expression, we employed the triplet loss. The triplet loss, Eq.~\eqref{equ:triplet}, forces the network to preserve the identity of the Source image. 
\begin{figure*}[!t]
		\centerline{\includegraphics[trim=50 0 50 0,width=13cm]{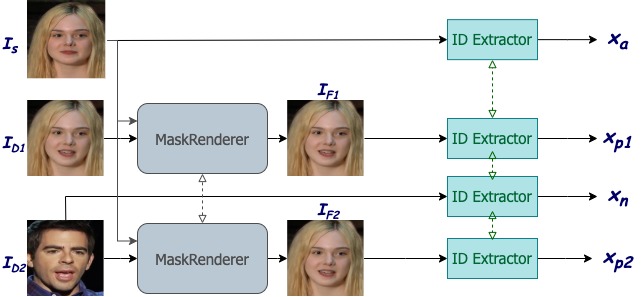}}
		\caption{Triplet loss illustration.}
		\label{fig:triplet}
\end{figure*}
\begin{equation}
    L_{T} = \sum_{a,p,n}^{}{[D_{x_a,x_p} - D_{x_a,x_n} + m ]_{+}} \text{; \thickspace\thickspace\thickspace } y_a = y_p \neq y_n
    \label{equ:triplet}
\end{equation}
Where $x_a$ is an anchor identity feature and $x_p$ and $x_n$ are positive and negative features, respectively. The input and reenacted images are passed through the pre-trained VGGFace2 \cite{cao2018vggface2} to extract identity features $x$. Pair $(x_a, x_p)$ belongs to the same identity class $y_a$, whereas negative samples have different identities $y_n$. The objective of triplet loss is to bring the reconstructed image's identity closer to the Source image while increasing its distance $D$ from a negative sample with the margin $m$. We expanded the triplet loss as Eq.~\eqref{equ:triplet2} to represent the cross-reenactment loss illustrated in Fig.~\ref{fig:triplet}. The second part of the equation captures difficult positive cases where Source and Driving have different facial shapes and poses.
\begin{equation}
    L_{T} = \frac{1}{|B|}\sum_{i=1}^{|B|}{[\alpha||x_{a} - x_{p_{1}}||_2 + (1 - \alpha)||x_a - x_{p_{2}}||_2 - ||x_a - x_{n}||_2 + m ]_{+}} 
    \label{equ:triplet2}
\end{equation}
A drawback of adding a triplet loss term to the network is that it requires additional computation. Moreover, triplet loss can cause negative bias and force the model to compromise the pose and expression to preserve identity. We address these issues by setting triplet loss to zero at initial epochs and applying it in the later epochs. Moreover, the weight between the first and second terms in the Eq.~\eqref{equ:triplet2} is tuned to reduce negative bias.

\textbf{\textit{Warping loss $L_W$:}} 
The warping loss attempts to bring the warped Source features closer to the Driving features $F_{Enc_{D}}$ to improve the warping accuracy and dense flow estimation. The warping loss is defined as:
\begin{equation}
    L_W = \sum_{i} mean({| \mathcal{T}(F_{Enc_{S, i}},\hat{S}(z)) -  F_{Enc_{D, i}}|}),
    \label{equ:warp}
\end{equation}
where the index $i$ represents the $i^{th}$ down-sampled layer in the generator. 
The next losses are defined as FOMM's \cite{siarohin2019first} loss functions.

\textbf{\textit{Equivariance losses $L_E$ \& $L_J$:}}
By adding the equivariance losses, we address the feature $F$ consistency issue to non-linear Thin Plate Spline (TPS) warping $\mathcal{T}_{TPS}$ \cite{siarohin2019first}. The $L_E$ and $L_J$ losses encourage the warped Driving features $\mathcal{T}_{TPS}(F_D)$ and motion $\mathcal{T}_{TPS}(J_D)$ to become closer to their corresponding values from the warped Driving frame $\mathcal{T}_{TPS}(I_D)$.

\textbf{\textit{Perceptual loss $L_P$:}}
The perceptual loss attempts to bring the reenacted image $I_F$ and Driving frame closer to each others using the pre-trained VGG\_19 network's features. In the following equations, the index $i$ represents the feature scale.
\begin{equation}
    L_P = \sum_{i} mean({| VGG_i(I_D) -  VGG_i(I_F)|})
    \label{equ:perceptual}
\end{equation}

\textbf{\textit{GAN loss $L_G$:}}
The MaskRenderer is trained similarly to a GAN, by alternating between the training of the generator and the discriminator. The discriminator consists of four layers for four feature maps, $F_M$, and a final discriminator map prediction, $D_M$.  
In the generator, the GAN loss $L_G$, defined as Eq.~\eqref{equ:GAN} attempts to make generated images look like real images. 
\begin{equation}
    L_G = \sum_{i} mean({( 1 -  D_{M_i}(I_F))^2})
    \label{equ:GAN}
\end{equation}
The \textbf{\textit{Feature Matching loss $L_F$}}, Eq.~\eqref{equ:feature_loss}, brings the Driving and reenacted frames' discriminator feature maps closer to each other. 
\begin{equation}
    L_F = \sum_{i}\sum_{j} mean({| F_{M_i,j}(I_D) -  F_{M_i,j}(I_F)|})
    \label{equ:feature_loss}
\end{equation}
 The index $j$ indicates the $F_M$ layer. At each iteration, first, the mentioned losses are combined as shown in Eq.~\eqref{equ:sum_loss} and passed to the optimizer.  
\begin{equation}
    L_{generator} = \alpha L_T + \beta (L_W + L_E + L_J + L_P + L_G + L_F) 
    \label{equ:sum_loss}
\end{equation}
Then, the discriminator is trained to discriminate between the real $I_D$ and generated $I_F$ photos to make the model robust to parameters that help distinguish between real and fake images. The discriminator loss $L_D$ is defined as follows: 
\begin{equation}
    L_D = \sum_{i} mean[{{( 1 -  D_{M_i}(I_D))}^2 + {D_{M_i}(I_F)}^2}]
    \label{equ:disc}
\end{equation}

\section{Experimental Setup}
\label{sec:Eval}
In this section, we discuss how we will train and evaluate MaskRenderer: the dataset, testing setup, and evaluation metrics. We also include implementation details for reproducability. 

\subsection{Dataset}
We used VoxCeleb1 dataset\footnote{\url{https://www.robots.ox.ac.uk/~vgg/data/voxceleb/vox1.html}}~\cite{nagrani2017voxceleb} to train and evaluate our model. It contains 22,496 videos from 1,251 identities that have been extracted from interview videos on YouTube. After preprocessing and cleaning the data (based on prior work~\cite{siarohin2019first}), we used 17,765 videos for training and 464 videos for evaluation. There is no overlap between the identities in the training and test sets.

\subsection{Inference stage}
Self-reenactment evaluations were performed on all 464 testing videos, where the first frame is set as the Source image, and the rest are the Driving images. On the other hand, in cross-reenactment evaluation, we randomly selected 61 test videos of 12 individuals. For each video, the first frame is assigned as the initial driving frame. This helps to consider the relative pose and expression distance between the initial frame and any other frame used as a driving image instead of the absolute pose value. Due to the difference in facial structures between Source and Driving, considering the relative changes in the Driving's face coordinates supports preserving the Source's identity. Additionally, The Source image is randomly selected from videos with different identities.

\subsection{Evaluation Metrics}
In the self-reenactment scenario, the final generated frame must replicate the Driving frame. Therefore, the Driving frame is the ground truth for pose, expression, and identity of the corresponding generated image. We considered standard metrics for comparing the SOTA models' self-reenactment performance \cite{ha2020marionette,wang2021safa,siarohin2019first}, as follows.
\textbf{L1 distance} calculates the mean absolute difference between Driving the generated frame. 
\textbf{Absolute Keypoint Difference (AKD)} measures the keypoints' contrast between Driving and the generated frame. \textbf{Absolute Identity Difference (AID)} measures the similarity between source and generated images' identity vector, estimated as in prior work \cite{feng2021learning}. Moreover, we computed the \textbf{Euclidean Pose Distance (EPD)} to compare the Driving and the generated frame's pose and expression using the pose embedding vectors \cite{deng2019accurate}.

For evaluating cross-reenactment performance, since there is no ground truth available when the Source and Driving people are different, we calculated cosine similarity between two features instead of calculating absolute distance. The cross-reenactment metrics are as follows. \textbf{Identity Similarity (ISIM)} determines how well the generated frame has preserved the Source's identity and facial structure. \textbf{Pose Similarity (PSIM)} and \textbf{Keypoint Similarity (KSIM)} quantify how close the poses and positional facial landmarks are, respectively, between the Driving and generated frames. 

The \textbf{Fréchet Inception Distance (FID)}  
estimates the difference between the normal distribution of the fake and real images' features by looking at the distance between the generated (fake) and Source (real) images, estimating how photo-realistic it is.
We used the pre-trained InceptionV3 for extracting images' features compared using FID as described in \cite{wang2021safa}.

\subsection{Implementation Details}

Our MaskRenderer was trained on the training data with 421 different identities. Both Driving and Source frames are randomly selected from a random patch of each person's videos. To cover various combinations of each identity's frames, the process is repeated 150 times with a batch size of 8 for 100 epochs. Both generator and discriminator learning rates are set to $2\times10^{-4}$ with $\beta_1 = 0.5$ and $\beta_2 = 0.99$. We used three NVIDIA 3090 GPUs with 24GB memory to train the model.
The loss hyper-parameters $\alpha$ and $\beta$ in the Eq. \ref{equ:sum_loss} are set to 10, and the triplet loss margin $m$ is set to 1. 
The facial features' number $K$ is set to 10 after performing experiments on the set $K=\{5, 10, 15, 32, 64\}$. Fewer points could not fully capture the face and body movements, while using more points resulted in the system becoming overly complicated (the model failed to accurately map and detect changes in the facial features). After testing the range of $[0.4,1]$ for $alpha$ in the Eq.~\eqref{equ:triplet2}, the higher weight, $\alpha = 0.8$, was selected to emphasize the first self-reenactment term and reduce the likelihood of any negative bias.
We replaced the nvdiffrast renderer in Deep3DFace \cite{deng2019accurate} with the Pytorch3D renderer\footnote{https://pytorch3d.readthedocs.io/en/latest/modules/renderer/index.html}, as it is computationally more efficient.

\section{Results}
\label{sec:Result}
In this section, we compare the MaskRenderer with state-of-the-art (SOTA) methods: FOMM~\cite{siarohin2019first}, SAFA~\cite{wang2021safa}, DaGAN~\cite{hong2022depth}, TPS~\cite{zhao2022thin}, and DualGen~\cite{hsu2022dual}. The metrics for SOTA are calculated for 
the released pre-trained models. 
All evaluations are performed on the same device. The inference time of MaskRenderer for cross-reenactment is less than 40 fps.

\subsection{Cross-Reenactment}
The main objective of this work is to improve the cross-reenactment performance when Driving and Source images have different facial structures and identities.  

\subsubsection{Quantitative Results}
Table~\ref{table:campare_cross} shows that MaskRenderer has the highest ISIM and KSIM scores. Higher identity and landmark similarity indicate that MaskRenderer better preserves the Source identity during pose and expression changes compared to the SOTA models. 
The visual qualitative comparison verifies this performance, especially for faces with very different structures. Moreover, MaskRenderer has the lowest FID score, confirming that our model generates the most photo-realistic images.
\begin{table}[ht]
    \caption{Cross-reenactment quantitative results. Bold values correspond to the best result among SOTA and our methods.}
    \begin{center}
    \resizebox{0.65\textwidth}{!}{\begin{tabular}{llllll}
            \toprule
            \toprule
            \textbf{Model}&\textbf{ISIM$\uparrow$}&\textbf{KSIM$\uparrow$}&\textbf{PSIM$\uparrow$}&\textbf{FID$\downarrow$}\\
            \midrule
            {FOMM\cite{siarohin2019first}}&0.873&0.899&0.512&59.522\\
            \midrule
            {SAFA~\cite{wang2021safa}}&0.856&0.902&\textbf{0.575}&55.456\\
            \midrule
            {DaGAN~\cite{hong2022depth}}&0.872&0.91&0.507&61.643\\
            \midrule
            {TPS~\cite{zhao2022thin}}& 0.851&0.91&0.571&69.144\\
            \midrule
            {MaskRenderer}&\textbf{0.891}&\textbf{0.914}&0.527&\textbf{49.469}\\
            \bottomrule
        \end{tabular}}
        \label{table:campare_cross}
    \end{center}
\end{table}

Although the pose similarity score of the SAFA and TPS is higher than MaskRenderer, these methods perform much worse in preserving the identity of Source image.  
The decrease in PSIM is because our method tries to make smooth transitions between frames, and in some cases, the generated result does not fully follow the Driving's pose in the area of the iris. However, this pose adjustment does not affect indicative features such as mouth movements, and the final result is highly realistic. All in all, MaskRenderer is smoother and more consistent in generating photo-realistic identity-preserved reenacted results than SOTA models.

\subsubsection{Qualitative Results} 
\begin{figure*}[!t]
\centering
\includegraphics[width=\textwidth,height=\textheight,keepaspectratio]{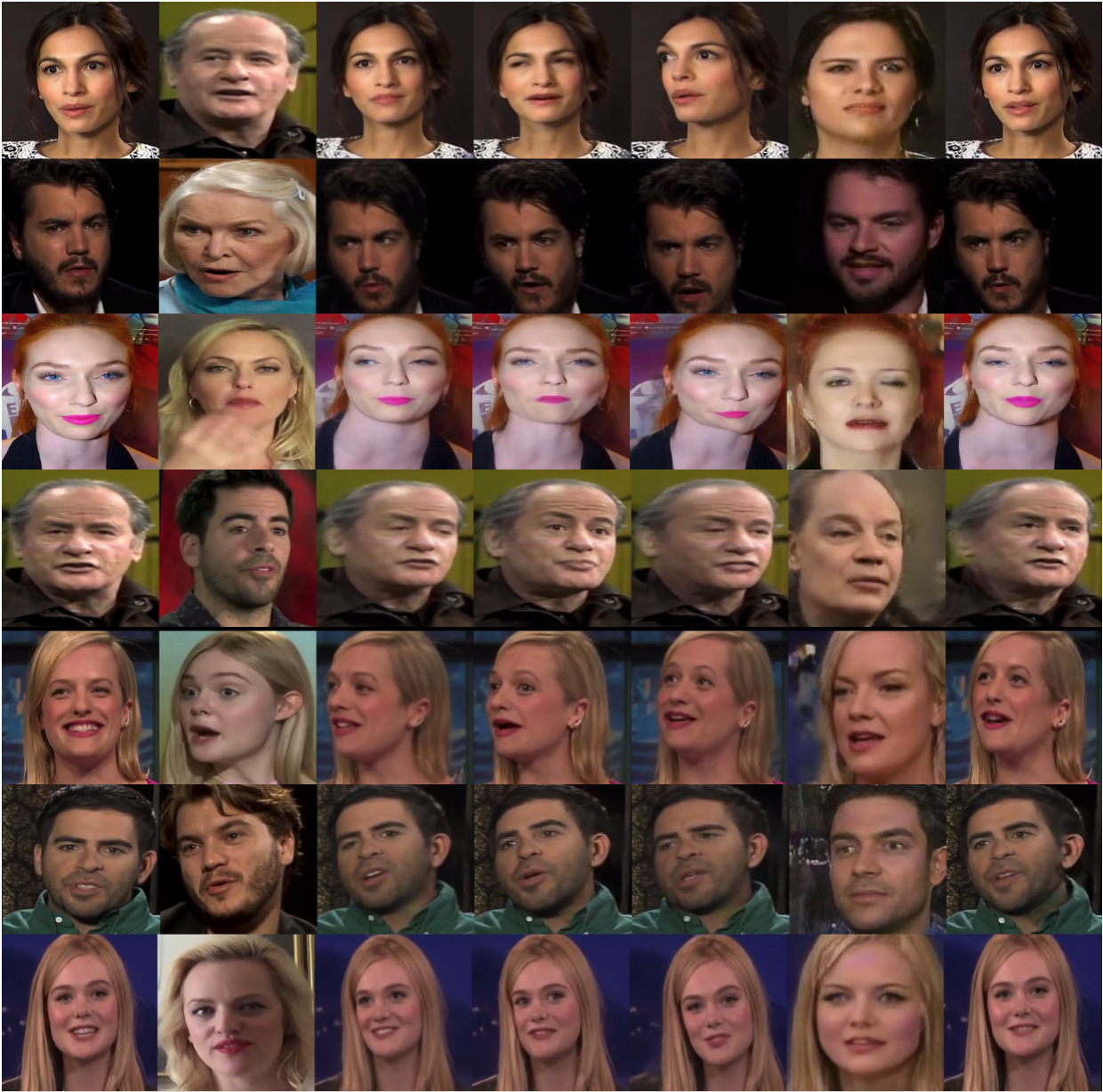} 
\begin{minipage}{.11\linewidth}
\centering
\caption*{Source}
\end{minipage}%
\begin{minipage}{.14\linewidth}
\centering
\caption*{Driving}
\end{minipage}
\begin{minipage}{.14\linewidth}
\centering
\caption*{FOMM}
\end{minipage}
\begin{minipage}{.14\linewidth}
\centering
\caption*{SAFA}
\end{minipage}
\begin{minipage}{.14\linewidth}
\centering
\caption*{DaGAN}
\end{minipage}
\begin{minipage}{.14\linewidth}
\centering
\caption*{DualGen}
\end{minipage}
\begin{minipage}{.11\linewidth}
\centering
\caption*{MaskRenderer}
\end{minipage}
\caption[Qualitative results]{Qualitative comparison of our method with the state of the art for cross-reenactment}
\label{fig:Qualitative}
\end{figure*}

Fig.~\ref{fig:Qualitative} shows the qualitative comparison of our MaskRenderer with the baseline (FOMM) and other SOTA models. The first and second columns show sample Source images and Driving frames, respectively. The rest of the columns represent the face reenactment results of each model corresponding to the Source and Driving images (first two columns). 
As shown in the last column, our method generates high-quality, high-fidelity, and photo-realistic face reenactment results. MaskRenderer visually outperforms SOTA methods in several aspects. First, MaskRenderer preserves the Source identity even when it is very different from the Driving face, while other models show signs of identity leakage more or less (as most apparent in the first row). Second, MaskRenderer handles the hand occlusion better than other models (see the third row). Third, MaskRenderer has the capability of generating reenactment in different head pose angles. Fourth, our method can generate fine-grained details, such as wrinkles, facial hair, and teeth. Fifth, it can also produce reasonable mouth movements, which is critical when the Driving person is speaking.

\subsection{Self-Reenactment}
\begin{table}[ht]
    \caption{Quantitative comparison of the proposed method with state-of-the-art methods for self-reenactment.}
    \begin{center}
    \resizebox{0.65\textwidth}{!}{\begin{tabular}{lllll}
            \toprule
            \toprule
            \textbf{Model}&\textbf{AKD}&\textbf{AID}&\textbf{$L_1$}&\textbf{EPD}\\
            \midrule
            {FOMM~\cite{siarohin2019first}}&1.378&0.12&0.045&0.917\\
            \midrule
            {SAFA~\cite{wang2021safa}}&\textbf{1.21}&0.118&0.041&\textbf{0.831}\\
            \midrule
            {DaGAN~\cite{hong2022depth}}&1.27&0.123&0.042&0.88\\ 
            \midrule
            {TPS~\cite{zhao2022thin}}&1.322&0.145&\textbf{0.039}&0.93\\
            \midrule
            {MaskRenderer}&1.395&\textbf{0.103}&0.044&1.023\\
            \bottomrule
        \end{tabular}}
        \label{table:campare_self}
    \end{center}
\end{table}

We also compared MaskRenderer with SOTA methods in self-reenactment cases where the Source and Driving frames are from the same person in a video. Table~\ref{table:campare_self} shows that our proposed method has the lowest AID, and the identity error has improved at least by 13\% compared to SOTA methods. This increase indicates that our method elevates identity preservation in self-reenactment scenarios and accurately reconstructs the facial details even for large pose changes, more than 45$^{\circ}$ head rotation. However, the AKD, EPD, and $L_1$ losses of MaskRenderer are higher than the previous methods. Adding the triplet loss helps reduce the bias of the same-identity reenactment during training. This modification comes with the trade-off of decreasing the self-reenactment's pose and landmarks accuracy, relative to SOTA methods. However, we believe the slight decrease in AKD and EPD are justified by the improved cross-identity image generation.

\section{Ablation Studies}
\label{sec:Ablation}

In this section, we conduct ablation studies to test the impact of adding the main components of our network to the baseline FOMM \cite{siarohin2019first} in cross-reenactment scenarios. 
Table~\ref{table:ablation-cross} quantitatively shows that adding the 3D parameters to the baseline (second row) helps mouth and body movements and pose changes; however, it does poorly in preserving the identity with high quality. On the other hand, including an attention unit to the hourglass network (third row) improves the ISIM and FID scores but leads to less accurate pose and expression generation. The forth row results validate the role of triplet loss in maintaining the Source identity while facing changes in pose and expression. This indicates the importance of considering identity change during training. The multi-scale occlusion masks and warping loss (fifth row) help with image background blending and inpainting unseen facial details. 

\begin{table}[th]
    \caption{Ablation on adding the main components of our network to the baseline for cross-reenactment}
    \begin{center}
    \resizebox{0.75\textwidth}{!}{\begin{tabular}{lllll}
            \toprule
            \toprule
            \textbf{Method}&\textbf{ISIM$\uparrow$}&\textbf{KSIM$\uparrow$}&\textbf{PSIM$\uparrow$}&\textbf{FID$\downarrow$}\\
            \midrule
            {FOMM (baseline)~\cite{siarohin2019first}}&0.873&0.899&0.512&59.522\\
            \midrule
            \midrule
            {w/ 3DMM}&0.871&0.902&\textbf{0.559}&52.339\\
            \midrule
            {w/ 3DMM+attention}&\textbf{0.926}&0.9&0.505&57.164\\
            \midrule
            {w/ 3DMM+Masks+$L_W$}&0.873&0.912&0.533&50.919\\
            \midrule
            {w/ 3DMM+$L_T$}&0.875&0.905&0.508&50.181\\
            \midrule
            {MaskRenderer}&{0.891}&\textbf{0.914}&0.527&\textbf{49.469}\\
            \bottomrule
        \end{tabular}}
        \label{table:ablation-cross}
    \end{center}
\end{table}

\begin{figure}[H]
\centering
\includegraphics[width=\textwidth,height=\textheight,keepaspectratio]{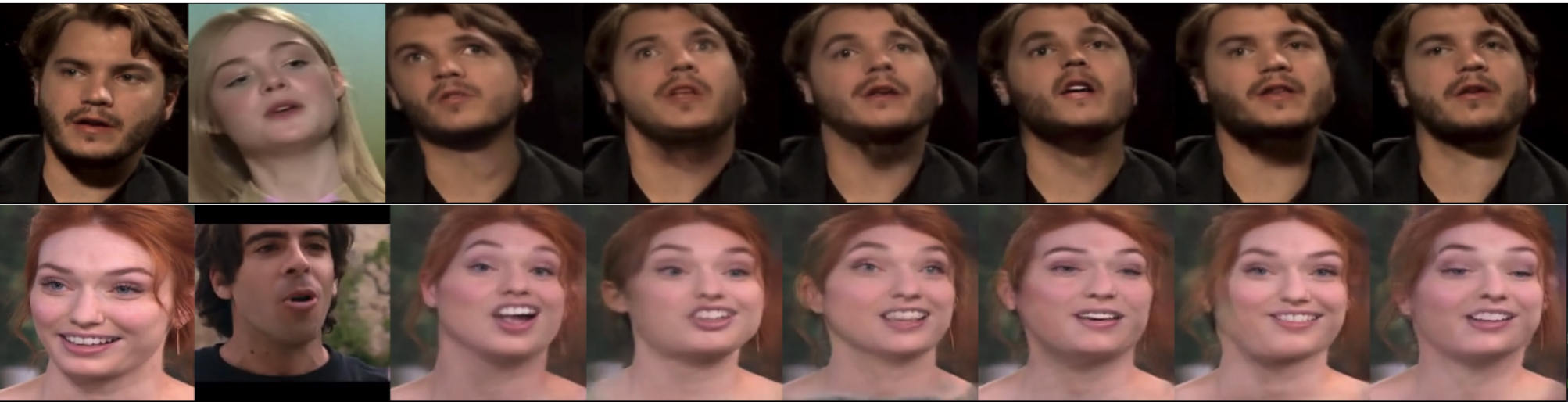} 
\begin{minipage}{.11\linewidth}
\centering
\caption*{\scriptsize Source}
\end{minipage}%
\begin{minipage}{.12\linewidth}
\centering
\caption*{\scriptsize Driving}
\end{minipage}
\begin{minipage}{.12\linewidth}
\centering
\caption*{\scriptsize Baseline}
\end{minipage}
\begin{minipage}{.12\linewidth}
\centering
\caption*{\scriptsize +3DMM}
\end{minipage}
\begin{minipage}{.12\linewidth}
\centering
\caption*{\scriptsize +3DMM+ Attention}
\end{minipage}
\begin{minipage}{.12\linewidth}
\centering
\caption*{\scriptsize +3DMM+ Masks+$L_W$}
\end{minipage}
\begin{minipage}{.12\linewidth}
\centering
\caption*{\scriptsize +3DMM+ $L_T$}
\end{minipage}
\begin{minipage}{.11\linewidth}
\centering
\caption*{\scriptsize MaskRenderer}
\end{minipage}
\caption[Qualitative ablation]{Qualitative ablation.}
\label{fig:ablation}
\end{figure}

From these quantitative results, we conclude that all the components in MaskRenderer contribute to the generation of highly realistic reenacted faces with high identity preservation. The final model has a high performance regarding identity, pose, and landmark, with the best FID score as compared to other models. Furthermore, MaskRenderer replicates face landmarks changes accurately and has the highest KSIM. Fig.~\ref{fig:ablation} also qualitatively supports all the aforementioned advantages of the proposed components. 

\section{Conclusion}
We have introduced MaskRenderer, a real-time identity-agnostic face reenactment framework that is robust to occlusion and large mismatches between Source and Driving facial structures. We incorporate the 3DMM model into the network to find a reliable 3D facial pose. The 3D and 2D motions are combined to precisely follow changes in the eyes and lips. By adopting the triplet loss, the impact of identity alteration is considered during training thereby reducing the bias to self-reenactment. 

We further designed multi-scale occlusion masks and the corresponding warping loss to improve the separation of foreground and background of a face and blending with the border. Ablation studies demonstrate the importance of each network's component in maintaining the facial structure and inpainting missing sections. 

Both qualitative and quantitative results illustrate that Maskrenderer improves the identity preservation of a Source image even for large poses compared to the SOTA models. The generated cross-reenacted results of the proposed method are photo-realistic, and the FID score is 10\% higher than the SOTA models. In the future, we plan to improve on the feature fusion and normalization in the generator to enhance the inpainting of the hair and teeth even further.
\label{sec:Conclude}

\section{Acknowledgment}
We wish to thank Mara Cairo, Deborah Akaniru, and Talat Iqbal Syed for their valuable expertise and support during this project. We would also like to thank Dr. Matt Taylor for his great advise and constructive feedback, which helped us to improve the quality of this work.
\label{sec:Acknowledgment}

\bibliographystyle{unsrt}
\bibliography{references}  

\end{document}